\documentclass[a4paper,fleqn]{cas-dc}

\pdfoutput=1

\usepackage[authoryear,longnamesfirst]{natbib}

\usepackage{tabularx,booktabs}
\usepackage{amsmath,amsfonts}
\usepackage{amssymb}
\usepackage{mathtools}
\usepackage{algorithmic}
\usepackage{algorithm}
\usepackage{array}
\usepackage[font=small,labelfont={bf}]{caption}
\usepackage{textcomp}
\usepackage{stfloats}
\usepackage{url}
\usepackage{verbatim}
\usepackage{graphicx}
\usepackage{multirow}
\usepackage{siunitx}
\usepackage{subfig}
\usepackage{lipsum}

\usepackage{natbib}
\setcitestyle{numbers,square}
\usepackage{bbm}
\usepackage{xcolor}
\usepackage{multirow}
\usepackage{makecell}
\usepackage{pifont}
\usepackage{bbding}
\usepackage{siunitx}
\usepackage{booktabs}
\usepackage{tabularx}
\definecolor{Green}{RGB}{0, 80, 0}
\definecolor{RevisedColor}{rgb}{0,0,0}
\begin{document}
\let\WriteBookmarks\relax
\def\floatpagepagefraction{1}
\def\textpagefraction{.001}

% \title[mode = title]{Visual Tracking for Generic Targets with Adaptive Detection Clustering and Tracklet-based Multi-hypothesis Association}
\title[mode = title]{Multi-tracklet Tracking for Generic Targets with Adaptive Detection Clustering}
% \title[mode = title]{Detectors knows better}

\tnotemark[1]

% Funding
\tnotetext[1]{
% The work presented here is included in the research project (RP/ESCA-03/2020) funded by Macao Polytechnic University, Macao, China.
    % This work was supported in part by the National Key Research and Development Program of China under Grant 2022YFB3306500, in part by the National Natural Science Foundation of China under Grant 61872025, in part by the Department of Faculty of Applied Sciences, Macao Polytechnic University, Macau, China, under Grant RP/ESCA-03/2020, and in part by HAWKEYE Group.
    This work was supported
    in part by the Science and Technology Development Fund, Macao SAR (File No. 0122/2023/AMJ), and
    in part by Macao Polytechnic University (File No. fca.f1a9.e38d.4).
    }

% Author
% style=chinese
\author[1,2]{Zewei Wu}[orcid=0000-0003-2289-0799]
\author[1]{Longhao Wang}[orcid=0009-0007-5111-1463]
\author[1]{Cui Wang}[orcid=0000-0001-9601-7824]
\cortext[1]{Corresponding author:cui.wang@mpu.edu.mo}
\cormark[1]
\author[2]{César Teixeira}[orcid=0000-0001-9396-1211]
\author[1]{Wei Ke}[orcid=0000-0003-0952-0961]
\author[3]{Zhang Xiong}[orcid=0000-0002-9421-1014]

\address[1]{Macao Polytechnic University, Macao SAR, China}
\address[2]{University of Coimbra, Coimbra 3004-531, Portugal}
\address[3]{Beihang University, Beijing 100191, China}

\begin{abstract}[S U M M A R Y]
    Tracking specific targets, such as pedestrians and vehicles, has been the focus of recent vision-based multitarget tracking studies.
    However, in some real-world scenarios, unseen categories often challenge existing methods due to low-confidence detections, weak motion and appearance constraints, and long-term occlusions.
    To address these issues, this article proposes a tracklet-enhanced tracker called Multi-Tracklet Tracking (MTT) that integrates flexible tracklet generation into a multi-tracklet association framework.
    This framework first adaptively clusters the detection results according to their short-term spatio-temporal correlation into robust tracklets and then estimates the best tracklet partitions using multiple clues, such as location and appearance over time to mitigate error propagation in long-term association.
    Finally, extensive experiments on the benchmark for generic multiple object tracking demonstrate the competitiveness of the proposed framework.
\end{abstract}

% Research highlights
% \begin{highlights}
% 	\item xxx
% \end{highlights}

\maketitle

% Main text
\section{Introduction} \label{sec:introduction}

\begin{figure}[t]
    \centering
    \includegraphics[width=8.5 cm]{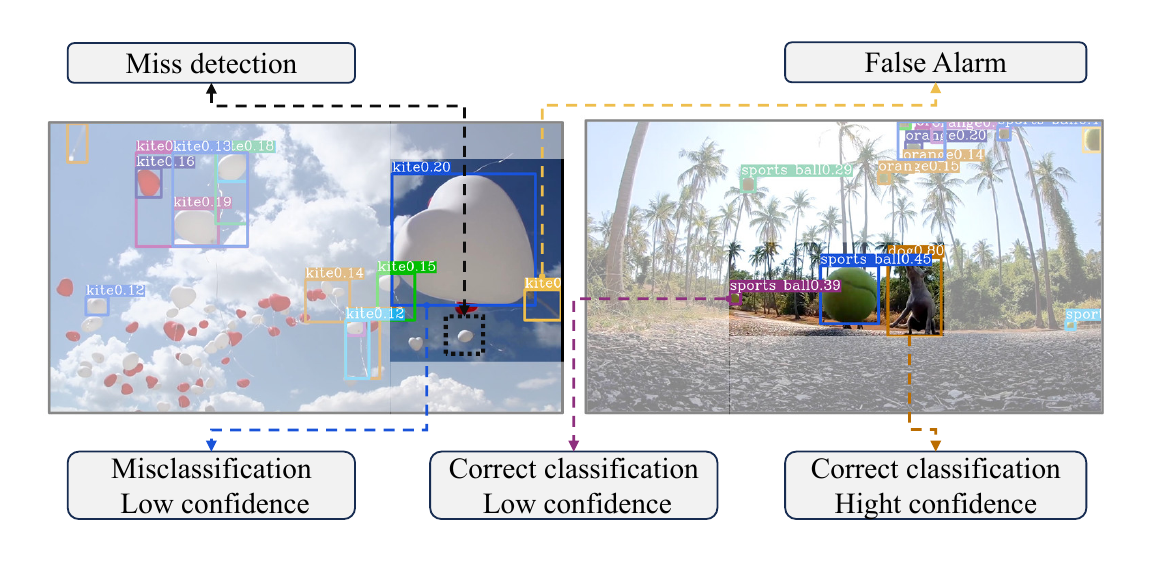}
    \caption{
        Detection failures occur in unseen categories.
        The results indicate that detectors are more confident in identifying targets with prior knowledge.
        For unseen targets, detectors may miss targets and produce false or low-confidence proposals.
        \label{fig:det_fails}
    }
\end{figure}

Multiple object tracking~(MOT) aims to identify targets and their trajectories within video sequences.
It requires not only accurately identifying the targets' positions and categories in individual images but also discovering temporal correlations of detections across timesteps.
Recent studies on MOT have revealed its immense potential in diverse applications, including smart cities~\cite{wu2022hybrid,wang2023blockchain,wang2022extendable,zhang2020multiplex}, biological research~\cite{wu2022online, wu2024online}, sport event~\cite{hu2024dctracker}, animal husbandry~\cite{kim2017thermal}, and the automation industry~\cite{zhang2023multi,zhang2023multiple}.
The breadth of these applications necessitates that tracking algorithms adapt to previously unencountered targets in deployment scenarios.
Such a requirement is encapsulated in a new task of visual tracking, known as Generic Multiple Object Tracking~(GMOT)~\cite{luo2013generic,luo2014bi,bai2021gmot}.
This task typically confronts target tracking with minimal a priori information.
Under the circumstance, object detectors often fail to measure accurately, resulting in increased unreliable measurements such as false alarms and missed detections, as shown in Fig~\ref{fig:det_fails}.
Consequently, those significant uncertainties lead to more relaxed constraints on motion and appearance which are the vital causes of fragmented tracking trajectories.

To address these issues, researchers have explored two solutions:
i) improving the generalization ability of detectors for generic targets, and
ii) enhancing the tracker's ability to recognize real trajectories amidst noise.
For the former, recent studies have demonstrated that open vocabulary detectors~\cite{li2022grounded, liu2023grounding, liu2024siamese} perform well in few-shot and even zero-shot detection scenarios.
These methods significantly enhance detector performance in general scenarios but require large-scale data and multimodal training.
For the latter, visual tracking methods typically rely on appearance features and Kalman filtering to remove noise.
Methods based on probabilistic tracking~\cite{rezatofighi2015joint,oh2009markov} are more reliable in complex environments.
However, these methods usually require more computational steps to improve tracking accuracy, thus reducing operational efficiency.
These two solutions are not conflicting but complementary.
This article focuses on the second solution.

To reduce measurement ambiguity, a more effective approach is to use short-term trajectories, known as~\emph{tracklets}, instead of individual detections.
Tracklet aggregats the target's average state over time which provides a more robust representation.
It is less susceptible to failed detections.
Tracklet-based tracking algorithms~\cite{wang2014tracklet,sheng2018iterative, zhang2020long} have garnered significant attention in pedestrian tracking.
Their core idea is to divide the entire video sequence into small subsequences of same length, cluster the detection results within each subsequence into tracklets, and further associating those tracklets.
However, in GMOT task, implementating an efficient tracklet generation solution poses significant challenges.
First, segmenting video sequence inevitably disrupts the completeness of tracks.
This interference is negligible when targets' motion is relatively stable but becomes problematic in dynamic scenarios.
Second, the failure of re-identification modules in GMOT task cannot be overlooked.
The absence of prior information about targets' appearance often results in re-identification modules failing to accurately retrieve targets through fine-grained features.
Consequently, trackers struggle to handle long-term occlusion events involving targets.

Trackers with delayed decision logic have proven effective in handling long-term occlusion events~\cite{blackman2004multiple,kim2015multiple}.
In these methods, the decision process for forming a trajectory is postponed until sufficient information has been collected.
Multiple Hypothesis Tracking~(MHT~\cite{rezatofighi2015joint,blackman2004multiple}) is a representative example.
MHT constructs track trees to indicate all possible trajectories and scores each branch to assess its credibility.
Inspired by this, we rethink tracklet generation and the multiple hypothesis association for the GMOT task and propose multi-tracklet tracking based on adaptive detection clustering.
Specifically, we introduce a new strategy for robust tracklet generation
This strategy takes into account scene dynamics such as the number of targets, target locations, and the confidence provided by the detectors when generating tracklets.
Based on this, our work also implements a corresponding tracklet-based representation and scoring mechanism to optimize performance.
Our approach improves the performance of the multi-hypothesis tracker in low-confidence detection scenarios.
Experimental results on benchmarks confirm the effectiveness of our method in the GMOT task.
% and offers insights for potential future parallel computation

The rest of the paper is organized as follows:
Section~\ref{sec:related_work} reviews related studies. Section~\ref{sec:preliminary} provides preliminary definitions.
Section~\ref{sec:methodology} outlines the proposed methodology and its key components.
Section~\ref{sec:experiment} describes the experimental setup, detailed factorial procedure, and presents the experimental results and analysis.
Section~\ref{sec:limitations} discusses the limitations of the proposed approach.
Finally, Section~\ref{sec:conclusions} summarizes this work and suggests directions for further study.

\section{Related work} \label{sec:related_work}
\subsection{Generic Multiple Object Tracking}
MOT typically relies on category-specific prior knowledge and models.
Although tracking algorithms~\cite{wojke2017simple,bewley2016simple,ciaparrone2020deep} were not initially designed for specific targets, their effectiveness is limited by the robustness of detection methods.
To address the generalization issue in MOT, the GMOT task was introduced in~\cite{luo2013generic,luo2014bi}.
GMOT involves recognizing and tracking objects in a video without predefined categories, presenting a more challenging task than MOT.

The significance of the GMOT task has recently been underscored with the introduction of the GMOT-40 dataset~\cite{bai2021gmot}.
This dataset enables performance testing of tracking algorithms across various categories and scenarios, offering a unified evaluation protocol based on the One-shot GMOT protocol, which ensures fair comparisons of different trackers. Consequently, there has been an increasing focus on GMOT.

Open Vocabulary Multi-Object Tracking~(OVMOT), exemplified by OVTrack~\cite{li2023ovtrack}, aims to track objects of arbitrary categories.
This approach enhances tracking performance by integrating visual language models (e.g., CLIP~\cite{radford2021learning}, GLIP~\cite{li2022grounded}) to improve the detection and feature representation of predefined categories.
Although effective, this method requires expensive and well-trained visual language models to bridge the domain gap between text and images.

Liu~\emph{et al}.~\cite{liu2024siamese} introduced Siamese-DETR, trained using only common detection training datasets.
This method employs a multi-scale query-based and simple Non-Maximum Suppression~(NMS) matching strategy~\cite{neubeck2006efficient}, achieving superior performance in the GMOT task compared to other methods.
Despite numerous studies on GMOT, the potential of data association strategies in this problem remains underexplored.
Our work aims to develop a tracklet-based tracking algorithm for the GMOT task, focusing on achieving reliable tracking under uncertain observations through robust tracklet representation and association strategy.

\subsection{Tracking with Tracklet}
Tracking algorithms based on tracklets~\cite{huang2008robust, roshan2012gmcp, sheng2018iterative, zhang2020long} typically replace discrete detections with tracklets as the fundamental tracking units, which are then merged to form complete trajectories.
These methods introduce a preprocessing stage for generating tracklets within the original tracking process.
Compared to discrete detections, tracklets incorporate an additional time dimension, enhancing target representation and effectively preventing error propagation and identity switching during tracking.

% Wang~\emph{et al}.~\cite{wang2016tracklet} proposed online feature learning and estimation for tracklet features.
% Shen~\emph{et al}.~\cite{shen2018tracklet} introduced a learning-based association network for bi-level tracklet association.
% Wang~\emph{et al}.~\cite{wang2019exploit} developed a multiscale TrackletNet for evaluating the similarity of tracklet pairs and clustering tracklets using appearance and temporal features.
% Chen~\emph{et al}.~\cite{chen2019aggregate} focused on the feature representation of track segments and introduced a multi-task convolutional neural network to incorporate spatio-temporal features into the modeling of short tracks' appearance.
% Zhang~\emph{et al}.~\cite{zhang2020long} addressed the issues of minimizing identity switching and creating extension problems for tracklets by proposing a deep model-based approach for long-term association with trusted tracklets.
Online feature learning and estimation for tracklet features were proposed by Wang~\emph{et al}.~\cite{wang2016tracklet}.
Shen~\emph{et al}.~\cite{shen2018tracklet} introduced a learning-based association network for bi-level tracklet association.
To evaluate the similarity of tracklet pairs and cluster them using appearance and temporal features, Wang~\emph{et al}.~\cite{wang2019exploit} developed a multiscale TrackletNet.
Chen~\emph{et al}.~\cite{chen2019aggregate} focused on the feature representation of track segments, introducing a multi-task convolutional neural network to incorporate spatio-temporal features into the modeling of short tracks' appearance.
Addressing the issues of minimizing identity switching and creating extension problems for tracklets, Zhang~\emph{et al}.~\cite{zhang2020long} proposed a deep model-based approach for long-term association with trusted tracklets.
Despite these advances, the potential of this approach for GMOT tasks remains underutilized.
Given the challenging nature of the GMOT task, this work focuses on flexibly generating reliable tracklets based on scenarios and accurately achieving long-term association.

\section{Concepts and Definitions}\label{sec:preliminary}

This section presents the foundational concepts and problem definitions pertinent to our study.

\subsection{Basic Definition}
Consider a video sequence composed of $N$ images, denoted as $\{X_1, X_2, \dots, X_N\}$.
A detection proposal is represented as $z=\left[\mathbf{b}, s\right]^T$, where the bounding box $\mathbf{b} =(x,y,w,h)$ is accompanied by a confidence score $s$.
Each detection proposal potentially signifies a true target.
The set of detections at the $t$-th frame is represented by $Z_t = \{z_{t,1}, z_{t,2}, \dots, z_{t,M}\}$, where $M$ denotes the total number of detections.
A tracklet, denoted as $\mathcal{Z}_{T,k}$, is a subset of detections over a brief interval $T = [t_m:t_n]$ with the same identity $k \in K$.
Each tracklet remains independent throughout its existence, ensuring that overlapping tracklets in time originate from different targets.

\subsection{Problem Formulation}
\begin{figure}[h]
    \centering
    \includegraphics[width=8 cm]{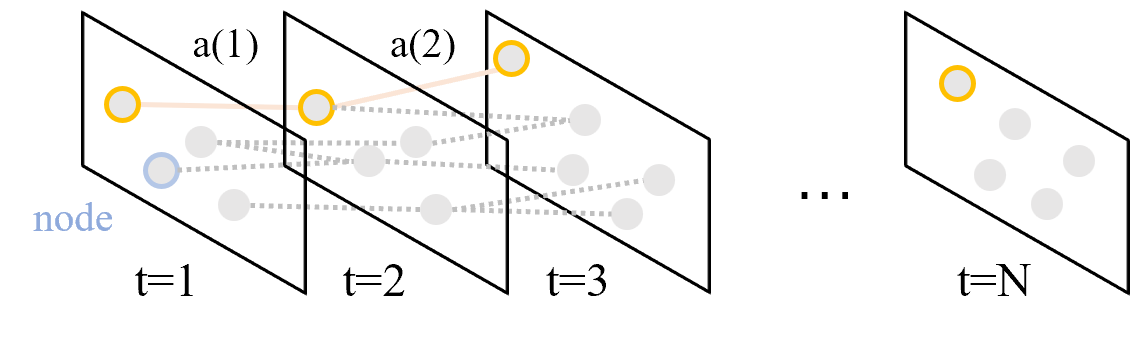}
    \caption{
        Illustration of a layered graph.
        \label{fig:layered_graph} % NEED TO IMPROVE
    }
\end{figure}

In our study, video data frames are conceptualized as a layered graph $G = (V,E)$, as shown in Fig.~\ref{fig:layered_graph}.
This graph consists of $N$ distinct layers of node sets $V$ and edge sets $E$.
For tracking applications, let $V$ represent the detection sets $\{Z_1, Z_2, \dots, Z_N\}$.
The edge sets $\{E_1, E_2, \dots, E_{N-1}\}$ denote links between detections in different frames.

Based on this, the goal of GMOT (similar to MOT) can be formulated as a multi-dimensional assignment problem as below,
\begin{equation}
    \begin{aligned}
        \min_{}     & \sum_{a \in \prod_{n=1}^{N-1} E_n}  c_a x_a                                                \\
        \emph{s.t.} & \sum_{a \in \prod_{n=1}^{t} E_n} x_{a} \leq \min (|Z_t|, |Z_{t+1}|), \text{for } t \in N-1 \\
                    & x_{a} \in\{0,1\}
    \end{aligned}
\end{equation}
where $a$ is an assignment candidate represented as a partition $\{a(1), a(2), a(3), \ldots, a(N-1)\}$.
In this context, $a(t)$ denotes a selection of paired detections $\{ (z_{t,i}, z_{t+1,j}) \mid z_{t,i} \in Z_t, z_{t+1,j} \in Z_{t+1} \}$.
$c_a$ is the cost of assignment and $x_a$ is a Boolean decision variable.
The constraints ensure that each target is assigned to at most one target in the subsequent frame.

Identifying the optimal assignment for $N > 2$ is an NP-hard combinatorial optimization problem.
This challenge is akin to the Maximum-Weight Independent Set (MWIS) or Set Packing problems.
Therefore, heuristic methods, along with significant computational effort, are generally employed to obtain good solutions.

\subsection{Tracklet in Multi-Hypothesis Framework}
The MHT algorithm effectively addresses the multidimensional assignment problem in tracking tasks~\cite{poore2006some}.
It enumerates all possible trajectories using a tree data structure, known as a track tree.
Each track tree represents multiple trajectory hypotheses for a target, with each leaf node corresponding to a detection proposal.
At each time step, the algorithm creates and updates these trees, evaluates them based on motion and appearance scores, and, when necessary, associates them across multiple frames to determine the best tracks.
Besides, the algorithm incorporates filtering mechanism and pruning scheme  to manage the size of the trees.
More details can be found in the paper~\cite{kim2015multiple}.
In our work, we utilize tracklets as the fundamental leaf unit rather than individual detections.
This subsection highlights the differences between the tracklet-based approach and the traditional MHT, focusing on tree creation and track scoring.

\begin{figure}[t]
    \includegraphics[width=8.5 cm]{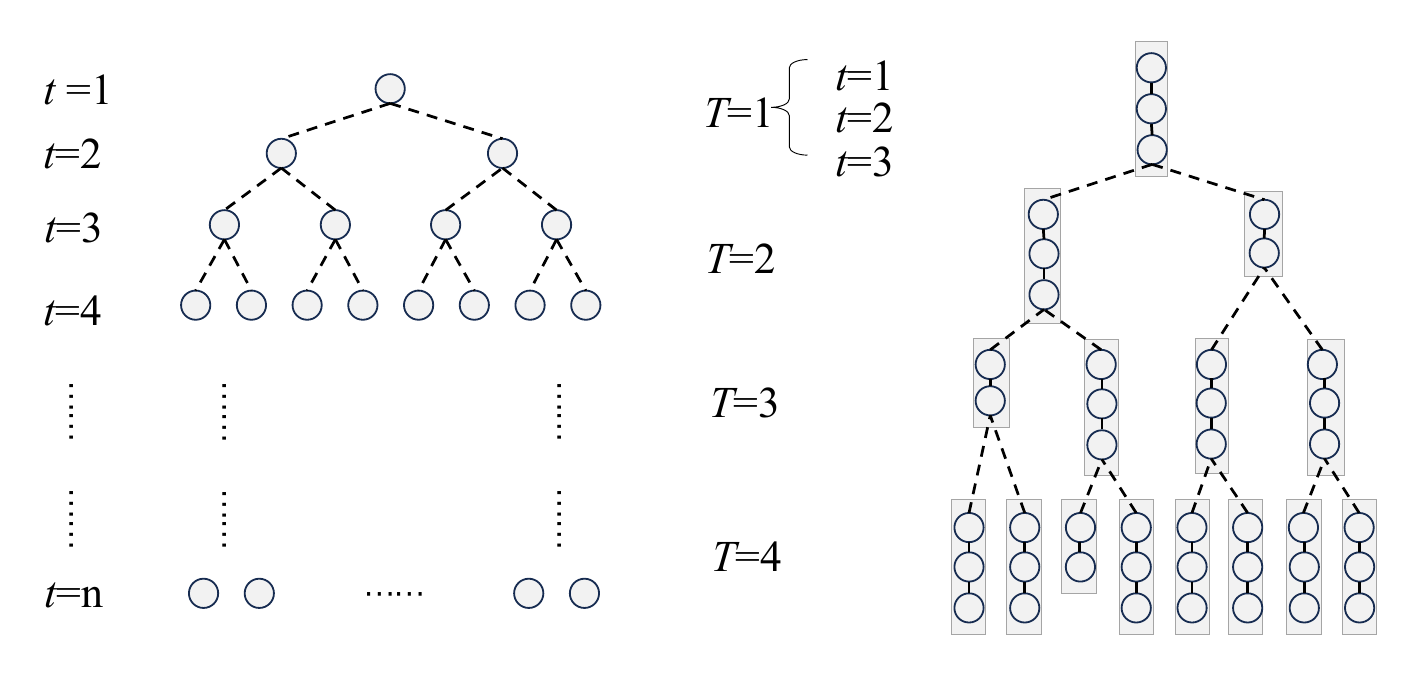}
    \caption{
        Illustration of differences between MHT and its tracklet-based version in tree creation.
        \label{fig:tree}
    }
\end{figure}

\paragraph{Tracklet Tree}
Initially, a track tree is created as the root node for each new detection, serving as the starting hypothesis set.
Detections from subsequent timesteps are added as leaf nodes, expanding the tree and forming new trajectory hypotheses.
All trees are updated over time until targets are lost, pruned, or the track ends.
To align a tracklet's expression with a detection concept, \emph{e.g.}, a tracklet with identity $k$, we represent it as a unique detection $z_{T,k}$, where $T = [t_m:t_n]$, defined as:
\begin{equation}\label{eq:det_align}
    \begin{aligned}
        z_{T,k} & = \mathcal{Z}_{T,k} = \{Z_{t,k}\}_{t=m}^{n} = \{(\mathbf{b}_{t,k}, s_{t,k})\}_{t=m}^{n}
    \end{aligned}
\end{equation}
On this basis, the tracklet-based approach is easily adapted to the original multi-hypothesis framework.
Fig.~\ref{fig:tree} illustrates this difference in tree creation.
% 或许需要描述更新过程

\paragraph{Tracklet Scoring}
In MHT, the likelihood function evaluates how well detector reports match the existing tracks.
Similarly, our goal is to evaluate the affinity between tracklets and existing tracks.
% maximum a posteriori (MAP) method
% Using established notations, this process can be represented as follows:
% \begin{equation}
%     \begin{aligned}
%         \mathcal{T}^* & = \arg \max _{\mathcal{T}} p \left(\mathcal{T} \mid \mathcal{Z}_a \right)                                 \\
%                     %   & = \arg \max _{\mathcal{T}_n } \prod  p\left(z_{T,i} \mid \mathcal{T}_n \right) p(\mathcal{T}_n)
%     \end{aligned}
% \end{equation}
% where $\mathcal{T}$ indicates a track set, $\mathcal{Z}_a$ is a set of tracklets, and $p \left(\mathcal{T} \mid \mathcal{Z}_a \right)$ is the posteriori probability of tracklet-track matching.
Tracking scores are typically calculated for each hypothesis based on observed information, such as the targets' location and appearance.
For a tracklet, the addition of the time dimension enhances the operability of track scoring.
The score of the $l$-th track at frame $t$ can be obtained using the following equation:
\begin{equation}
    S^{\text{track}}_{t,l}= \sum_i  w_{\text{i}} \cdot S_{\text{i}}(z_{T,k},\mathcal{T}_l)
\end{equation}
where $S_{\text{i}}(\cdot, \cdot)$ and $w_{\text{i}}$ denote one of the scoring items and the corresponding weight, respectively.
This usually involves the motion score and appearance score.
For a more detailed calculation of tracklet scoring, see Section~\ref{subsec:tracklet_scoring}.

\subsection{Tracklet Generation}\label{tracklet_generation}
\begin{figure}[t]
    \includegraphics[width=8.5 cm]{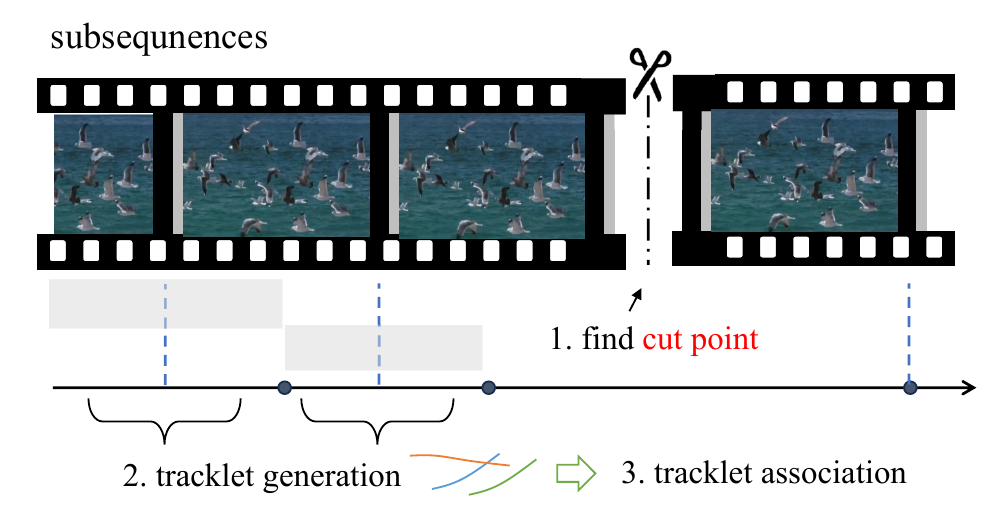}
    \caption{
        Illustration of tracklet-based tracking.
        \label{fig:tracklet-based_tracking}
    }
\end{figure}

In tracklet-based tracking methods, as illustrated in Fig.~\ref{fig:tracklet-based_tracking}, a video sequence is first divided into subsequences.
Within these subsequences, all detections are clustered into several partitions.
Each partition, representing a set of detections with identical labels, forms the initial tracklets.
The accuracy of these initial tracklets is essential for subsequent tracking.

\paragraph{Sequence Partitioning}
Existing approaches typically utilize hyperparameters of length $L$ to generate subsequences of uniform length.
These approaches can be categorized into two strategies: direct slicing and slicing with a sliding window.

The first strategy splits the entire video sequence into $n=\lceil \frac{N}{L} \rceil$ parts based on the subsequence length $L$.
The subsequences can be represented as:
\begin{equation}
    \begin{aligned}
        \left\{X_1, X_2, \ldots, X_L\right\},\left\{X_{L+1}, X_{L+2}, \ldots,X_{2L}\right\}, \ldots, \\ \left\{X_{\lceil \frac{N}{L} \rceil L+1}, X_{\lceil \frac{N}{L} \rceil L+2}, \ldots, X_{N}\right\}
    \end{aligned}
\end{equation}
This strategy is straightforward and easy to implement; however, it has notable disadvantages.
It disrupts the correlation within the video sequence, leading to incorrect links between detections due to incomplete information.

To address this issue, the window rolling approach, also called sliding window, is proposed.
It moves one fixed step $S$ at a time, where $S \leq L$, typically with $S$ equal to 1.
This method results in $N-L+1$ subsequences, which can be expressed as:
\begin{equation}
    \begin{aligned}
        \left\{X_1, X_2, \ldots, X_L\right\},\left\{X_{1+S}, X_{2+S}, \ldots, X_{L+S}\right\}, \\
        \ldots, \left\{X_{N-L+1}, X_{N-L+2}, \ldots, X_N\right\}
    \end{aligned}
\end{equation}
This approach preserves as much complete information as possible but it products more subsequence which incurs additional computational costs.

In summary, both approaches have their own drawbacks.
In our work, a trade-off strategy is adopted to split a more reasonable subsequence (see Section~\ref{sec:sequence_partitioning}).

\paragraph{Subsequence Clustering}
After sequence partitioning, we consider clustering the detections within each subsequence.
The clustering problem for subsequences is typically regarded as a small-scale MOT problem, which can be addressed using various data association methods, such as the nearest-neighbor method, bipartite matching, and others.
To fully utilize the correlation between any two frames within a subsequence, some works~\cite{sheng2018iterative, zhang2020long} model the problem as a weighted K-bipartite graph and implement detection clustering via integer linear programming.
This approach maximizes the use of inter-frame information.
However, due to the introduction of the NP-Hard problem, the time overhead becomes unacceptable as the subsequence size increases.
How to make more efficient utilization of inter-frame information of subsequences without excessive computational overhead is the focus of our work.
Our proposed solution is presented in Section~\ref{subsec:density_based_clustering}.

% There are various types of solutions such as greedy algorithm, bipartite graph matching, network flow maximization, subgraph clustering

\section{Methodology}\label{sec:methodology}

In this article, we propose a multi-tracklet tracking method with adaptive detection clustering, termed MTT.
Our approach emphasizes generating reliable tracklets and establishing associations between them.
To achieve this, we introduce a robust sequence partitioning method and a density-based detection clustering strategy, ensuring flexible and robust tracklet generation.
Subsequently, tracklet operations are seamlessly integrated into a multi-hypothesis association framework, leveraging the short-term benefits of tracklet representation and the long-term advantages of multi-hypothesis tracking to address the challenges of the GMOT task.

\begin{figure*}[t]
    \centering
    \includegraphics[width=16 cm]{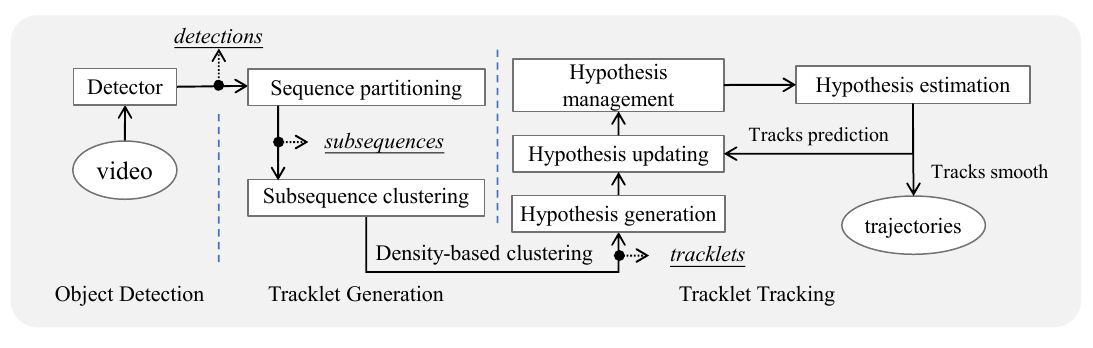}
    \caption{
        Flow diagram of the proposed multi-tracklet tracking.
        \label{fig:workflow}
    }
\end{figure*}

\subsection{Overview}
As previously discussed, the primary challenges in the GMOT task arise from detection noise and the degradation of motion and appearance features. Inspired by tracklet-based methods, we propose a multi-tracklet tracking framework with adaptive detection clustering to overcome these issues, as illustrated in Fig.~\ref{fig:workflow}.

\subsection{Tracklet Generation with Adaptive Detection Clustering}
% insight：把目标检测器作为一种抽象的传感器，检测指标则被看作传感器产生的测量，一组视频帧中的检测结果即为多维度的时间序列。
Tracklet generation is a crucial step in tracklet-based algorithms, as the quality of the generated segments often determines the accuracy of subsequent tracking.
This subsection explores how sequence slicing can be performed to minimize additional impacts.
First, we need to reach a consensus that "detectors know your scenarios." This implies that the detector's reports can reflect changes in the tracking scenario's dynamics, such as the number of detected targets, the confidence scores of detections, and their change rate in neighboring frames.
% Such variations are common in the detection process.
For example, occlusion events can lead to a sudden increase in low-confidence detections.
Zhang~\emph{et al}.~\cite{zhang2022bytetrack} and Wang~\emph{et al}.~\cite{wang2022extendable} mentioned a similar phenomenon.
Another situation where the number of targets can change abruptly is when multiple targets appear or disappear from the scene.
Recognizing these changes helps us to segment the sequence more effectively, as discussed in Section~\ref{sec:sequence_partitioning}.

When performing partitioning operation, it is natural to consider whether this operation can reduce the problem size or be used for parallel calculation.
Although parallelism is possible, this work focuses on the former.
As mentioned in Section~\ref{tracklet_generation}, the process of generating tracklets involves solving an NP-Hard problem, and the problem size affects its solvability.
To our knowledge, this has not been investigated in previous work.
It is feasible to reduce the computational cost by keeping the subsequence length small, such as 2 frames.
However, this strategy has two expected drawbacks: first, despite the small slices, there may be many targets within a single image frame; second, it will undoubtedly weaken the representation capability of tracklets.
Therefore, we propose a new partitioning method, which breaks the above limitation and allows detections in even one frame to be segmented into different parts.
We provide the solution in Section~\ref{subsec:density_based_clustering}.

\subsubsection{Robust Sequence Partitioning}\label{sec:sequence_partitioning}
% 使用了所有的包含低置信度检测的数据来过滤，后续通过NMS=0.5来过滤
Given a video sequence $\mathcal{X} = \{X_1, X_2, \dots, X_N\}$, we have a set of detection counts $\mathcal{M} = \{|Z_1|, |Z_2|, \dots, |Z_N|\}$, denoted as $\{M_1, M_2, \dots, M_N\}$, indexed by $\{1, 2, \dots, N\}$.
Note that all detections, including low-confidence ones, are considered to construct a kind of count curve.
We define a gradient metric of the detection count curve as $r_t = \frac{|M_t - M_{t+\Delta}|}{\Delta}$ with $\Delta=1$.
Generally, this gradient metric changes smoothly over time, except in specific cases such as occlusion, false detection, and sudden changes in the number of targets.
These cases commonly lead to matching errors in tracking.
We aim to find these time points through the detection information.
These points divide the video sequence $\mathcal{X}$ into a series of subsequences, denoted as $\mathcal{X}_{\text{sub}} = \{ \mathcal{X}_{\text{sub}}(1), \mathcal{X}_{\text{sub}}(2), \ldots \}$.

We propose a straightforward method to minimize the impact of sequence partitioning on tracking performance.
First, we apply one-dimensional median filtering to the curve to avoid spike noise, obtaining $\mathcal{M}_{\text{filtered}}$ using a filter window parameter $w_\text{median}$.
Next, we set an upper threshold $d$ to determine whether a time $t$ is a reasonable cutting point.
For the initial $i = 0$, during the loop $t = 1, 2, 3, \ldots, N$, for each $X_t \in \mathcal{M}_{\text{filtered}}$, follow the expression below:
\begin{equation} \label{eq:simple_split}
    \mathcal{X}_{\text{sub}}(i) = \begin{cases} \mathcal{X}_{\text{sub}}(i) \cup X_t, & \text{if } r_t < d; \\
              i=i+1, \mathcal{X}_{\text{sub}}(i) \cup X_t, & \text{otherwise} \end{cases}
\end{equation}

Although the aforementioned strategy is intuitive, it is insufficient for achieving robust sequence partitioning.
We must identify two types of change trends in the curve: (i) a faster change trend, indicative of fast dynamic events such as short-term occlusion, and (ii) a gradual change trend, reflective of significant alterations in the actual number of objects in the scene.
Our goal is for the segmented subsequence to encompass the first type of changes while minimizing the inclusion of the second type.
This method allows short-term occlusion to be managed within the subsequence, thereby reducing the risk of propagating matching errors.
Long-term changes can be assessed globally through multi-tracklet association.
We first define a condition as:
\begin{equation}\label{eq:cond1}
    (M_t - \min \left(\mathcal{X}_{\text{sub}}(i)\right) < d) \text{ or } \\
    (\max \left(\mathcal{X}_{\text{sub}}(i)\right) - M_t < d)
\end{equation}
Therefore, Eq.~\ref{eq:simple_split} can be extended to the following form:
\begin{equation}
    \mathcal{X}_{\text{sub}}(i) = \begin{cases}
        \mathcal{X}_{\text{sub}}(i) \cup X_t, & \text{if Eq.~\ref{eq:cond1} is true}; \\
        i = i + 1, \mathcal{X}_{\text{sub}}(i) \cup X_t, & \text{otherwise}
    \end{cases}
\end{equation}

\begin{figure}[t]
    \centering
    \includegraphics[width=7.5cm]{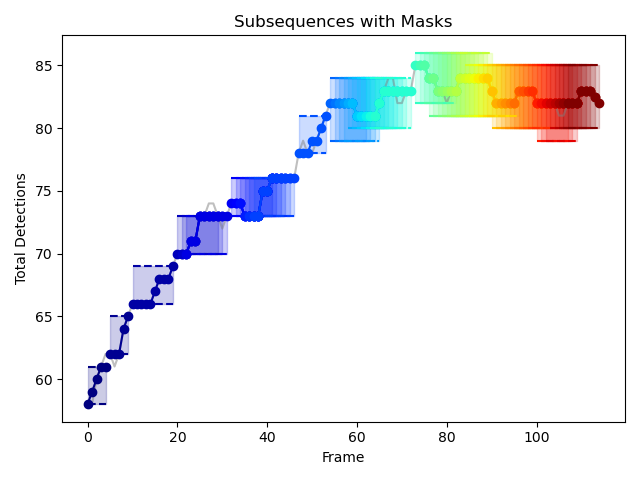}
    \caption{
        Flexible sequence partitioning based on detector reports.
        \label{fig:vsp}
    }
\end{figure}

In addition, we aim to develop an adaptive mechanism to maintain a stable computational load.
For excessively long subsequences, we implement variable sliding window settings.
We define a maximum sliding window size $l_{\text{max}}$ and an upper limit $u$ for the total number of detections in a subsequence.
If $\sum_{j=1}^{l_{\text{max}}} \mathcal{X}_{\text{sub}}(i)[j] > u$, then $l_{\text{max}}$ is reduced until $l_{\text{max}}$ equals 1 or the condition is satisfied.
% If the change threshold is set very high, causing the entire sequence to contain only one subsequence, this method will degenerate into the ordinary sliding window method.
An example of intuitive subsequence segmentation is shown in Fig.~\ref{fig:vsp}.

\subsubsection{Density-based Tracklet Generation}\label{subsec:density_based_clustering}
% 滑动窗口带来的冗余和轨迹冲突会进行相应的滤除
% However, it is important to note that simply cutting the entire picture into several equal parts will destroy the frame image's integrity.
% To achieve computational efficiency,
% It should be noted that reducing the length of the subsequence to lessen the computational burden is effective, but it significantly reduces the available time dimension.
Our subsequence partitioning method comprehensively considers both the length of the subsequence and the number of detections within it, balancing computational efficiency and accuracy.
Given the potentially excessive number of targets in a scene, the computational cost can exceed expectations even with a single image frame.
Therefore, it is prudent to explore segmentation methods at the subframe level.
To our knowledge, this has not been discussed in previous work.
A straightforward solution is to divide the image into several equal parts.
However, this naive approach may pose greater challenges to target tracking, especially when targets are unevenly distributed in the image.
In response, we propose a more reasonable method of subframe-level segmentation by density.
This is because targets are limited by their motion capabilities and will not move far from their current positions in a short period.
Therefore, the feasibility of matching distant targets is greatly reduced.
Eliminating these matching candidates will increase speed without affecting matching accuracy.

For a subsequence $\mathcal{X}_{\text{sub}}(i)$, we can obtain all detections, represented as $\{ Z_n \cup Z_{n+1} \cup \ldots \cup Z_m \}$.
Note that these detections require NMS~\cite{neubeck2006efficient} and confidence threshold filtering to ensure reliable results.
We can easily obtain a set of these detection center points, denoted as $Z_{\text{sub}}$.
Given a radius $\epsilon$, the $\epsilon$-neighborhood is defined as
\begin{equation}
    N_\epsilon(p) = \{q \in Z_{\text{sub}} \mid \operatorname{dist}(p, q) \leq \epsilon\}
\end{equation}
If $|N_\epsilon(p)| \geq \delta$, then $p$ is a core point and forms a cluster.
For implementation, we set parameters ($\epsilon, \delta$) and provide the distance matrix of $Z_{\text{sub}}$, then perform clustering using the DBSCAN algorithm.
In our work, the distance matrix is as follows:
\begin{equation}
    \left(\begin{array}{cccc}
            \text{dist}\left(z_1, z_1\right) & \text{dist}\left(z_1, z_2\right) & \cdots & \text{dist}\left(z_1, z_n\right) \\
            \text{dist}\left(z_2, z_1\right) & \text{dist}\left(z_2, z_2\right) & \cdots & \text{dist}\left(z_2, z_n\right) \\
            \vdots                           & \vdots                           & \ddots & \vdots                           \\
            \text{dist}\left(z_n, z_1\right) & \text{dist}\left(z_n, z_2\right) & \cdots & \text{dist}\left(z_n, z_n\right)
        \end{array}\right)
\end{equation}
where $\text{dist}(\cdot,\cdot)$ indicates the Euclidean distance between two detections.

\begin{figure}[t]
    \centering
    \includegraphics[width=8.5cm]{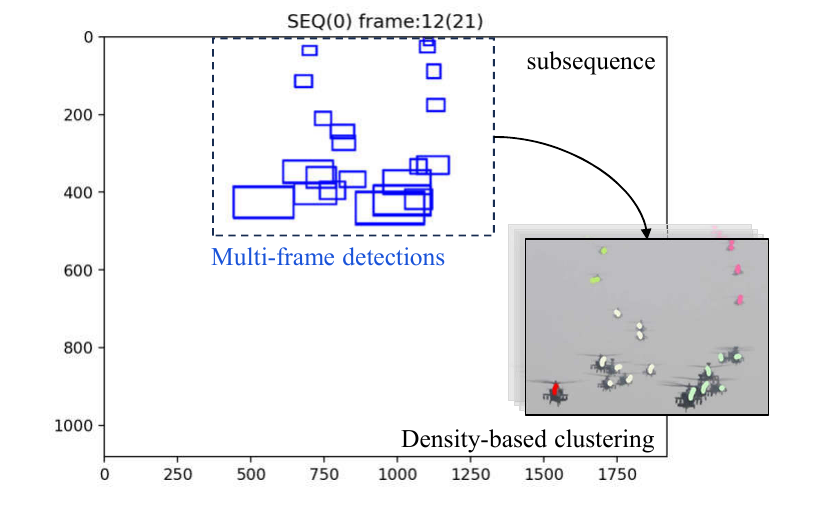}
    \caption{
        Density-based clustering in subsequence partitioning.
        \label{fig:dbscan}
    }
\end{figure}

Similarly, we can perform clustering using a multi-weighted normalized distance matrix instead of directly using pixel distance for density-based clustering.
This approach requires redefining the distance formula.
An example using a detection pair $(z_1, z_2)$ is:
\begin{equation}
    \text{dist}(z_1, z_2) =
         \alpha \cdot \cos(\phi(z_1), \phi(z_2)) + \beta \cdot \|z_1 - z_2\|_2
\end{equation}
where $\cos(\phi(z_1), \phi(z_2))$ represents their appearance feature cosine distance, and $\|z_1 - z_2\|_2$ represents their location distance.
Moreover, $(\alpha, \beta)$ are pair weighting coefficients, and $\phi(\cdot)$ denotes a feature extractor.
This allows us to perform density-based clustering through a similarity radius in an abstract sense, providing better adaptability for different scenarios.
Both methods have similar effects on the GMOT task, with pixel distance offering a more intuitive physical interpretation.
A example of clustering is shown in Figure~\ref{fig:dbscan}.

% 如过是边跟踪边分割的话，可以考虑用上次卡尔曼滤波的最大位置偏差来进行聚类

After obtaining the preliminary clusters, we construct a $K$-partite graph $G$ from each cluster to address the tracklet generation problem.
Let $G = (V, E, W)$, where the node set $V$ represents all observations in the cluster.
The edges in $E$ are undirected and carry a weight $w \in W$, representing the affinity between the two linked nodes, expressed as:
\begin{equation}
\begin{aligned}
    w_{ij}    & = \prod_{k=1}^5 w_{k, ij}                          \\
    w_{1, ij} & = \left\|z_i - z_j\right\|_2 \\
    w_{2, ij} & = \cos \left(\phi(z_i), \phi(z_j)\right) \\
    w_{3, ij} & = \mathbbm{1}_{ w_{1, ij} < \text{min}(z_i[\text{width}],z_j[\text{width}]) \cdot |\text{fr}(i) - \text{fr}(j)| } \\
    w_{4, ij} & = \mathbbm{1}_{w_{2, ij} \ge \theta_{\text{app}}} \\
    w_{5, ij} & = \mathbbm{1}_{\text{fr}(i) \neq \text{fr}(j)}
\end{aligned}
\end{equation}
Here, $z[\text{width}]$ represent the width of detetion $z$.
The function $\text{fr}(\cdot)$ retrieves the time index of detections.
$w_{1}$ and $w_{2}$ describe the affinity of linked nodes, while $w_{3}$, $w_{4}$, and $w_{5}$ provide filtering conditions to eliminate unreasonable candidates.
Position and similarity constraints set thresholds for appearance similarity and location distance.
Based on this, we can model tracklet generation as an optimization problem:
\begin{equation}
\begin{aligned}
    \arg \max _{e_{ij}} & \sum_{i, j} w_{ij} e_{ij}     \\
    \mbox{\em s.t. }       & e_{ij} + e_{jk} \leq e_{ik} + 1 \\
                       & e_{ij} \in \{0,1\}  \\
                       & e_{ij} = e_{ji}
\end{aligned}
\end{equation}
where $e_{ij}$ is a Boolean variable, allowing the optimization problem to be solved using integer linear programming.
The first constraint ensures that all nodes in any cluster in the solution are connected to each other.
Thus, we can generate the tracklets by sloving this optimization problem.

%     % \item Trajectory Smoothing: Smoothing of the target trajectory to reduce noise and error. Commonly used methods include Kalman smoothing and particle filtering.
%     % \item Trajectory Prediction: Based on the current trajectory information, the future position of the target is predicted. This is important for early action or warning.
% \end{itemize}

\subsection{Tracklet Tracking}

\begin{figure}[t]
    \includegraphics[width=8.5 cm]{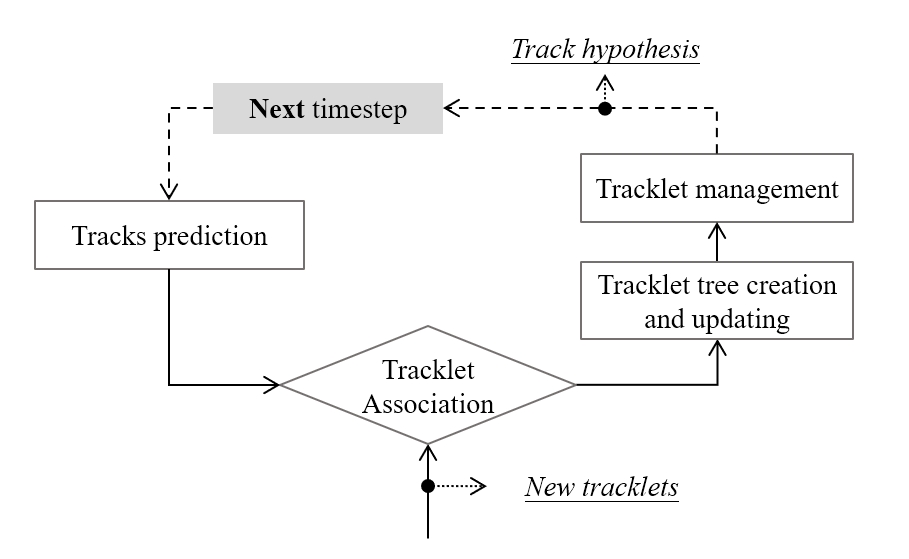}
    \caption{
        Illustration of the procedure of tracklet tracking.
        \label{fig:flow}
    }
\end{figure}

Unlike the traditional HMT framework, our approach emphasizes the establishment and linking of tracklets.
This brings three clear benefits:
i) leaf nodes contain more feature information and are easily distinguishable;
ii) it effectively avoids generating an excessive number of hypotheses that are difficult to manage;
iii) it maintains a longer tolerance window for pruning, thereby reducing decision uncertainty.
To integrate tracklet processing into the original MHT framework, we align tracklet expression with traditional detection through the definition in Eq.~\ref{eq:det_align}.
However, we must still design operations for tracklet representation and scoring.
As shown in Fig.~\ref{fig:tree}, tracklets replace detections as leaf nodes in the track tree.
Thus, during tree construction, we need to measure the affinity between tracklets and associate them.
As illustrated in Fig.~\ref{fig:flow}, tracklet tracking after adaptive detection clustering involves the following steps:
\begin{enumerate}[i)]
    \item \textbf{Hypothesis Generation}: For each timestep, multiple tracklet trees are generated based on current new tracklets.
    \item \textbf{Hypothesis Updating}: New tracklets are associated with existing track hypotheses, involving the tracklet scoring process.
    \item \textbf{Hypothesis Management}: This involves managing and maintaining all track hypotheses, including merging similar trajectories and terminating implausible ones, to retain only the most likely hypotheses and reduce computational complexity.
    \item \textbf{Hypothesis Estimation}: State estimation for each hypothesis is performed using filters~(e.g., Kalman filter) to predict the target's future position and velocity.

\end{enumerate}

\subsubsection{Tracklet Tree Creation and Updating}
% Fig.~\ref{fig:workflow}
% Since many discrete errors generated by the detector are filtered out during the tracklet generation phase, the tracking hypothesis based on tracklets is more reliable and reduces the impact of false detections.
This phase involves two main steps: constructing new trees and iteratively updating existing trees.
After dividing the subsequence, new trees are constructed using generated tracklets as root nodes.
Following the first timestep, new trees for the tracklets are built, and existing hypothesis trees are extended using current tracklets.
To ensure computational efficiency, we perform a preliminary screening of tracklets, filtering out those with low affinity to the parent node using a gating mechanism.
We utilize tracklet features, such as position and appearance over time.

To maintain computational efficiency, we employ appearance and motion constraints to control the increase in tree size.
We assume that a target's features do not change abruptly over a short period.
Therefore, we define the constraint of changed location with a Kalman filter to estimate changes in its motion state:
\begin{equation}\label{eq:motion_constraint}
    d_{\mathrm{motion}} = \left(\hat{{z}}_{T,k} - {z}_{T,k}\right)^{\top} \left(\Sigma_{T,k}\right)^{-1} \left(\hat{{z}}_{T,k} - {z}_{T,k}\right) \leq \theta_{mot}
\end{equation}
where $\hat{{z}}_{T,k}$ is the predicted motion state and $\Sigma_{T,k}$ is the covariance term of innovation.
Moreover, we use a weighted average according to the detection confidence score $z[s]$ to represent the appearance feature of the tracklet:
\begin{align}\label{eq:app_constraint}
    \phi (z_{T,k}) = \sum_i \frac{\operatorname{z}_i[s]}{\sum_i(\operatorname{z}_i[s])} \cdot \phi({z}_i) \\
    d_{\text{app}} = \cos (\phi(z_{T_i,p}),\phi(z_{T_j,q})) \leq \theta_{app}
\end{align}
In addition to the standard leaf nodes, we utilize a \emph{dummy} node from the MHT framework to manage target loss state.
This dummy node inherits its parent node's features, such as appearance and position, and is iteratively updated within a tolerance window.
If the tolerance time is exceeded, the target is deemed lost, and the corresponding trajectory is terminated.

\subsubsection{Tracklet Scoring}\label{subsec:tracklet_scoring}

Each branch in the tracklet tree represents a potential trajectory suggestion. Calculating the plausibility of each suggestion through a scoring function is crucial for tracklet association.
Our goal is to accurately evaluate hypothetical trajectories using the available information and construct reasonable features to characterize the target.
Our approach leverages the rich information of tracklets to provide an additional advantage in target feature construction.
It consists of three components: motion, appearance, and detection confidence scores.
Let's take the current tracklet hypothesis ${z}_{T,k}$ as an example.
For motion scoring, similar to Kim's formulation~\cite{kim2015multiple}, we use the log-likelihood of location observation.
The final form is defined as the ratio between the tracklet hypothesis and the null hypothesis (combining Eq.~\ref{eq:motion_constraint}):
\begin{align}
    & S_{\mathrm{mot}}(k) = \ln \frac{V_\text{space}}{2 \pi} - \frac{1}{2} \ln \left|\Sigma_{T,k}\right| - \frac{d_{motion}}{2}
\end{align}
where $V_\text{space}$ represents the observation space. We assume the target motion follows a Gaussian distribution.
For appearance scoring, given a default appearance score of the null hypothesis $\theta_{null}$, we obtain the log-likelihood ratio of appearance observation (combining Eq.~\ref{eq:app_constraint}) as:
\begin{align}
    & S_{\mathrm{app}}(k) = -\ln \left(1 + e^{-2 d_{app}}\right) - \ln \theta_{null}
\end{align}
Additionally, we consider the confidence scores of the tracklet and propose confidence scoring:
\begin{align}
    & S_{\mathrm{conf}}(k) = \tanh (\bar{z}_{T,k}[s] - \theta_s)
\end{align}
where $\bar{z}_{T,k}[s]$ is the average confidence score of detections in tracklet ${z}_{T,k}$.
And $\theta_s$ gives a lower bound on the confidence score, below which it is converted to the (0,$-\infty$) interval and vice versa.

Combining the above, the total score is calculated as:
\begin{equation}
    S_{\text{total}}=w_{\mathrm{mot}} S_{\mathrm{mot}}(k) + w_{\mathrm{app}} S_{\mathrm{app}}(k) + w_{\mathrm{conf}} S_{\mathrm{conf}}
\end{equation}
where $w_{\mathrm{mot}}$, $w_{\mathrm{app}}$, and $w_{\mathrm{conf}}$ are the corresponding weight terms. If the score is greater than 0, the tracklet hypothesis is more likely to have originated from a real target.

\subsubsection{Global Tracklet Association}
% Conflicts arise when a trajectory segment is assigned to multiple trajectory trees, which means that different trees may represent the same trajectory.
% To resolve these conflicts, a set of optimal or suboptimal hypotheses must be selected from multiple candidate hypotheses to maximize a certain evaluation criterion. To select the best trajectory hypothesis from the tree structure, we transform it into a weighted graph model. Global trajectory selection can be described as a global optimization problem. In this model, the weight of each node represents the quality of the corresponding observation, and the weight of each edge represents the reliability of the data association between nodes. We solve the MWIS problem to determine the largest set of non-intersecting high-weight nodes as the best trajectory hypothesis. The MWIS algorithm can find an approximately optimal independent set in polynomial time, which is a set of trajectory hypotheses in which no two nodes share the same observation. The input of the MWIS algorithm is a weighted graph, where each node represents a trajectory hypothesis, each edge represents a potential conflict between two hypotheses, and the weight of each node reflects the score of the hypothesis.
% Our goal is to find the optimal solution of the graph formed by all trajectory trees. After obtaining the optimal solution, a pruning strategy similar to is adopted. Except for the selected branches, all other branches are pruned, so that only one node survives at the end of a frame.

\begin{figure}[h]
    \includegraphics[width=8.5 cm]{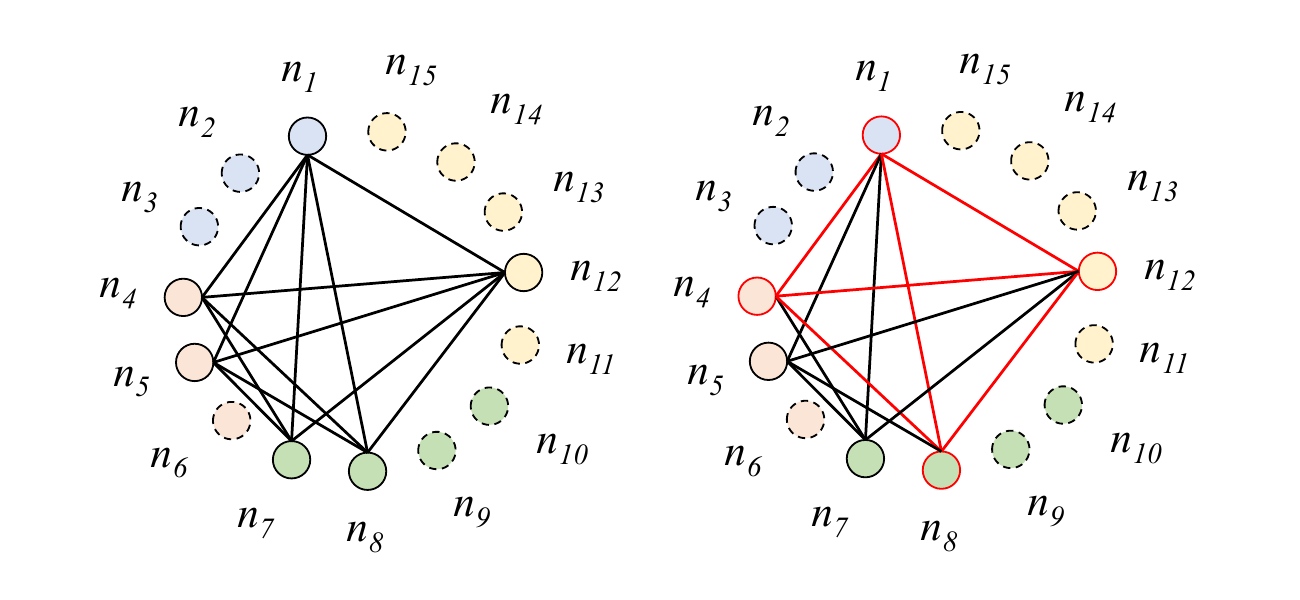}
    \caption{
        Illustration of weighted undirected graph for global tracklet association.
        \label{fig:weighted_undirected_graph}
    }
\end{figure}

Conflicts arise when a tracklet is assigned to multiple trees, indicating that different tracklet trees may represent the same target's trajectory.
To resolve these conflicts, the best hypothesis must be selected from multiple candidates through global tracklet association.
Following the formulation~\cite{papageorgiou2009maximum}, this problem can be transformed into a weighted graph model $G=(V, E, W)$, described as the MWIS problem, as shown in Fig.~\ref{fig:weighted_undirected_graph}.

MWIS identifies the `good' subset of disjoint nodes (vertices) in the graph, representing track hypotheses where no two nodes share the same observations.
In this model, each node represents a tracklet hypothesis, each edge signifies a potential conflict between two hypotheses, and the weight of each node reflects the hypothesis's score.
The objective is to determine the largest set of high-weight, disjoint nodes as the best trajectory hypothesis.
This process is represented by the following expression:
\begin{equation}
    \begin{aligned}
        \arg \max _{x_i} \sum_{i=1} w_i x_i\\
        \emph{s.t.}\ x_i + x_j \leqslant 1, \forall (x_i, x_j) \in E \\
        x_i \in \{0,1\}
    \end{aligned}
\end{equation}
where $x$ is a Boolean variable that determines whether the $i$-th tracklet belongs to the optimal solution.

Our goal is to find the optimal solution for the graph consisting of all tracklet trees.
After obtaining the optimal solution, we employ the N-Scan pruning strategy to ensure only one selected branch node remains at the end of the frame.
For the identified global trajectories, further post-processing is required to repair the broken trajectories.
In our work, the Kalman filter is chosen as a smoother for the trajectory reports.

% 还有轨迹段补全和平滑

% \subsubsection{Optimization}

% \paragraph{Prune}
% Exist three points (in original MHT framework) need to improved for efficiency and accuracy:
% i) each new tracklet is created as a new tree;
% ii) n-SCAN pruning starts slowly;
% iii) each tree maintained the same depth of pruning.

% % 即使剪枝深度只设置为2，也会产生几何质变，导致内存消耗快速增加 ...

% \begin{figure}[h]
%     \includegraphics[width=8.5 cm]{44.png}
%     \caption{
%         This is a figure about pruning. place holder
%         \label{fig44}
%     }
% \end{figure}

% To reduce computational complexity and memory usage, it is necessary to periodically filter and prune the trajectory trees. Filtering involves eliminating hypotheses with scores below a certain threshold, thereby reducing unreliable assumptions. Typically, a gating function based on the Mahalanobis distance is employed to discard observations that do not match the predicted state, thus minimizing the likelihood of mismatches. Pruning entails retaining only the optimal or suboptimal paths within each trajectory tree to decrease redundant hypotheses. Additionally, pruning limits the number of branches per node, thereby controlling the size of the hypothesis space. This paper implements the k-scan algorithm for this purpose. Finally, a strategy based on probability and lifespan is utilized to remove low-quality or outdated trajectory hypotheses, thereby freeing memory resources.

% \paragraph{Post-progress}
% xxx interpolation, filter, Kalman filter smooth ...
\begin{table}[t]
    \centering
    \caption{Main parameters in use.
    }
    \label{tab:params}
    \renewcommand\arraystretch{1.3}
    \setlength\tabcolsep{2.5pt}
    \begin{tabular}{ll}
    \hline
        Parameter & Value \\ \hline
        resolution of input video   & $1920 \times 1080$   \\ \hline
        \multicolumn{2}{l}{Tracklet Generation}\\  \hline
        filter window ($w_\text{median}$) & 5 \\ \hline
        confidence score threshold ($\theta_s$) & 0.1 \\ \hline
        NMS threshold  & 0.5 \\ \hline
        upper threshold of the gradient metric ($d$) & 5  \\ \hline
        step size of sliding window ($S$)  & 1  \\ \hline
        max length of subwindow ($l_{max}$)  & 5  \\ \hline
        max detection count in subsequence ($u$)  & 70  \\ \hline
        DBSCAN parameters ($\epsilon,\delta$) & (80,2)  \\ \hline
        % cluster parameters ($\alpha,\beta$) & (0.9,0.1)  \\ \hline
        motion distance threshold ($\theta_{mot}$) & 15  \\ \hline
        appearance distance threshold ($\theta_{app}$) & 0.85  \\ \hline
        \multicolumn{2}{l}{Tracklet Tracking}\\  \hline
        pruning depth ($k$) & 2  \\ \hline
        appearance likelihood of null hypothesis ($\theta_{null}$) &  0.3 \\ \hline
        max patience window of target missing  &  10 \\ \hline
        max number of leaves &  8 \\ \hline
        scoring weights  ($w_{\mathrm{mot}},w_{\mathrm{app}},w_{\mathrm{conf}}$) &  (0.1,0.9,3.0) \\ \hline
        % \multicolumn{2}{l}{Environments}\\  \hline
        % Encoder (pretrained on ImageNet) &  ResNet-18  \\ \hline
        % programging language &  Python 3.9 \\ \hline
        % neuro library
        % &  Pytorch 1.9~\cite{paszke2019pytorch} \\ \hline
        % operation systerm
        % &  Win11 \\ \hline
        % CPU
        % &  AMD R7-4800S  \\ \hline
        % GPU
        % &  RTX2060 Max-Q  \\ \hline

    \end{tabular}
\end{table}

\section{Results}\label{sec:experiment}
In this section, we compare and analyze the effectiveness of our proposed multi-tracklet tracking on the GMOT task.
We present both qualitative and quantitative tracking results on the public benchmark GMOT-40~\cite{bai2021gmot}.
Our method is implemented in Python 3.9, with the main hardware configuration including an AMD 4800S CPU and an RTX 2060MQ GPU.
We employed the ResNet-18~\cite{he2016deep} encoder (pretrained on ImageNet~\cite{deng2009imagenet}) as the feature extractor.
It runs in the PyTorch 1.9~\cite{paszke2019pytorch} runtime library.
More details are shown in Table~\ref{tab:params}.

\subsection{Benchmark}
\paragraph{Dataset}
We focus on the GMOT task and evaluate our method using the GMOT-40 benchmark~\cite{bai2021gmot}.
Its dataset is a pioneering public resource for GMOT, encapsulating numerous real-world tracking challenges such as occlusion events, object entry and exit, motion blur, and appearance deformation.
It comprises 40 annotated sequences evenly distributed across 10 object categories.
The collection features a wide variety of objects, with sequences showing significant differences within the same category.
Additionally, the objects are densely packed in scenarios, with dozens of the same type potentially appearing in a single frame.
This benchmark provides an excellent testbed for evaluating the performance of tracking algorithms on GMOT task.

\paragraph{MOT protocols}
We utilized common MOT metrics~\cite{bernardin2008evaluating} including Multi-Object Tracking Accuracy~(MOTA $\uparrow$), IDentity switch~(IDs $\downarrow$), False Positive~(FP $\downarrow$), False Negative~(FN $\downarrow$), and ID F1 score~(IDF1~\cite{ristani2016performance} $\uparrow$).
MOTA integrates IDs, FP, and FN across all frames, thus providing a comprehensive accuracy score.
This reflects the tracker's proficiency in consistent tracking without target loss or the generation of false tracks.
The IDF1 assesses the tracker's capability in maintaining consistent identity assignments over time, computed as the harmonic mean of ID precision and ID recall.

% 40G 的内存 AMD R7-4800S RTX2060MQ
% 如果窗口内的节点数不加限制，会超出内存 LP_solver
% 如何树的节点树不加限制，也会超出内存... MHT node
% 设定内存上限和时间限制
% 这里要注意win_size=1的时候可能是tracklet=2？
\begin{table*}[t]
    \caption{Performance and time-cost of tracklet generation.
        % 可以预见的是，窗口越大效果应该越好
    } \label{tab:tracklet}
    \begin{tabular}{c|c|cc|cc|cc|c}
        \toprule
        Method                   & max size & tracklets & tracklets* & IDs   & IDs* & time (s) & time* (s) & pass rate (\%) \\
        % \midrule
        % & &  &   &  &   &   &    &              \\
        \midrule
                                 & 1           & 92725     & 20969        & - & -  & -      & -      & -             \\
        \midrule
        \multirow{8}*{Fix-length window}     & 2           & 49694     & 11093        & 41269 & 8437   & 266      & 73      & 100             \\
                                 & 3           & 49128     & 11006        & 41124 & 8411   & 486      & 83      & 100             \\
                                 & 4           & 47438     & 10980        & 39572 & 8395   & 487      & 76      & 97.5         \\
                                 & 5           & 44659     & 10945        & 37326 & 8372   & 1293     & 150     & 92.5         \\
                                 & 6           & 37772     & 10915        & 31505 & 8352   & 1550     & 159     & 82.5         \\
                                 & 7           & 31062     & 10881        & 25464 & 8327   & 1705     & 155     & 65          \\
                                 & 8           & 23422     & 10849        & 18820 & 8311   & 1912     & 156     & 52.5         \\
                                 & 9           & 14319     & 10815        & 11327 & 8285   & 687      & 600     & 35          \\
                                 & 10          & 10774     & 10774        & 8259  & 8259   & 845      & 845     & 30           \\
        \midrule
        % 1  & 92725         & 72.50      & 383.67s    & 1        \\
        \multirow{8}*{Variable-length window}  & 2           & 50677     & 11201        & 42011 & 8514   & 757      & 174     & 100             \\
                                 & 3           & 49195     & 10974        & 41269 & 8377   & 787      & 185     & 100             \\
                                 & 4           & 48655     & 10904        & 40601 & 8325   & 738      & 168     & 100             \\
                                 & 5           & 48408     & 10834        & 40401 & 8279   & 753      & 170     & 100             \\
                                 & 6           & 48178     & 10780        & 40211 & 8239   & 783      & 176     & 100             \\
                                 & 7           & 48074     & 10737        & 40112 & 8201   & 792      & 179     & 100             \\
                                %  & 8           & 47064     & 10702        & 40060 & 8181   & 735      & 172     & 97.5         \\
                                 & 8           & 47987     & 10702        & 40060 & 8181   & 750      & 172     & 100         \\
                                 & 9           & 47927     & 10657        & 39992 & 8138   & 785      & 179     & 100             \\
                                 & 10          & 47879     & 10638        & 39960 & 8125   & 781      & 180     & 100             \\
        \bottomrule
    \end{tabular}

    \noindent{\footnotesize{* indicates statistics of valid 12 sequences (30\% of total 40) for fairness comparasion.}}
\end{table*}

\paragraph{GMOT protocols}
To comprehensively evaluate our method, we introduce multiple GMOT evaluation protocols, summarizing one-shot detection, zero-shot detection, and normal detection.
These GMOT protocols provide different sources of detection results.
The detection in the one-shot protocol is obtained using a single-object tracking method by iteratively retrieving the template features of the initial frame. These detection results come from the GlobalTrack tracker and are provided in the original work~\cite{bai2021gmot}.
The results of the zero-shot detection protocol usually come from object detectors based on vision-language models.
In our work, we chose GLIP~\cite{li2022grounded} as the zero-shot detector. Finally, the normal detection protocol refers to using object detection models trained on common object detection data.
Since most popular detectors, such as YOLOv5~\cite{glenn_jocher_2021_5563715}, perform poorly on GMOT, we chose DETR-Siamese~\cite{liu2024siamese} with a SwinT~\cite{liu2021swin} backbone pretrained on the COCO dataset~\cite{lin2014microsoft}.

% This protocol accepts the bounding box of a single target object in the initial frame and aims to detect and track all objects of the same category. The process is facilitated by two main components: one-shot detection and multi-object association.
% The one-shot detection technique is developed by leveraging the single object tracking (SOT) method. This method labels the object by extracting the appearance features of the object of that category from the initial frame. The location of the object in subsequent frames is then searched by the SOT method based on the label. All the results retrieved in this way are regarded as the results of one-shot detection on the object, and this detection process is performed in an iterative manner to accumulate the results of all objects.
% In addition to the one-shot detection protocol, we also use the zero-shot detection protocol. The results of the zero-shot detection protocol are derived from the detection results of existing zero-shot detectors (GLIP, DETR-Siamese) on the GMOT data machine.

% \paragraph{Baselines}
% xxx...xxx...

\subsection{Tracklet Effectiveness Analysis}

Our work focuses on flexible tracklet generation and object tracking based on tracklet association.
In this subsection, we demonstrate the effectiveness of our approach through comparative analysis in terms of tracklet accuracy, tracklet representation capability, and tracklet association efficiency, highlighting the enhancements in accuracy and robustness.

\subsubsection{Effect of Tracklet in Accuracy and Timecost}

\begin{figure}[t]
    % 另外一边y轴可以放准确率
    % \centering 33.84 / 14.54 2789
    % 1-10:[4594.28, 2813.62, 2809.91, 2769.76, 2738.66, 2843.23, 2842.88, 2789, 2738.72, 2722.18]
    \includegraphics[width=8.5 cm]{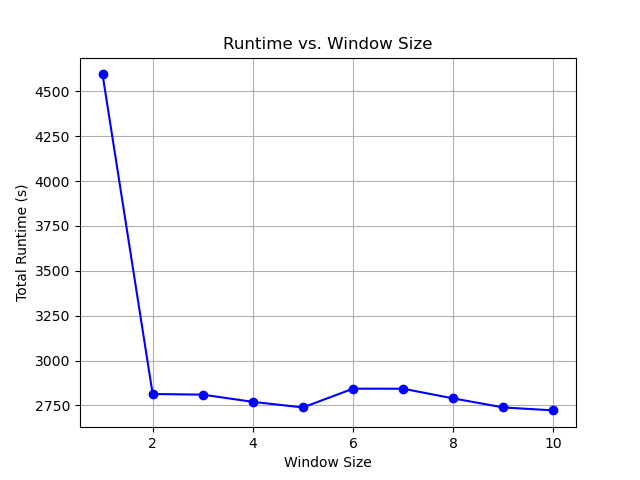}
    \caption{
        Illustration of the relationship between tracking runtime and window size.
        %我们的方法具有更好的可伸缩性，不会随着窗口扩大而明显增加xxx
        \label{fig:track_time}
    }
\end{figure}

To evaluate the effectiveness of tracklet operations in our method, we conducted experiments detailed in Table~\ref{tab:tracklet}.
We compared fixed-length window tracklet generation with variable-length window tracklet generation, both of which are modifications of our original approach.
We assessed the impact of window length on the number of tracklets, tracklet IDs, time taken, and the algorithm's pass rate.
The pass rate indicates the proportion of successfully solved sequences out of 40 as the problem size increases.
As discussed in Section~\ref{tracklet_generation}, heuristic methods are employed during tracklet generation.
Without effective control of problem size, the optimization problem may fail due to resource constraints. Therefore, it is crucial to consider the impact of problem size on the algorithm's execution.
Another aspect that needs explanation is the items marked with an asterisk~(*).
For a fair comparison, those items indicate that we only compare the sequences generated by tracklets that passed all tests.
This represents 30\% of the total sequences, which is actually 12 sequences.

We present the scenario where the window size is set to 1, indicating that the tracklet method is not utilized.
In this case, the number of tracklets equals the number of valid detections, which is significantly higher than in other methods.
This implies that the subsequent tracking process requires the most association calculations.
%%%%%%%%%%%%%%%%%%%
In the fixed-length window experimental group, an intuitive correlation can be observed: as the window size increases, the total number of tracklets and IDs decreases, while the time required for generation increases significantly.
Specifically, as the tracklet length increases, the algorithm's pass rate decreases.
This indicates that as the window size grows, the generated tracklets become longer and more accurate, but the problem size also increases, making it more prone to failure.

In the variable-length window experimental group, we observed similar phenomena, such as changes in the total number of tracklets and IDs.
However, it is noteworthy that while accuracy increased, the required computation time did not change significantly, and the algorithm maintained a 100\% pass rate.
Under fair comparison settings, the time spent by the former in the window size range of 5 to 8 did not change significantly.
This is because the time-consuming tasks failed, and the remaining tasks had a slight increase in time consumption, resulting in no significant change.
Similarly, from the actual time spent, the time for the former increased from 1293 seconds to 1912 seconds when the window size was set to 5-8, but the pass rate decreased from 92.5\% to 52.5\%.
The time spent on failed sequences is not included in the total time.

Overall, the variable-length window method shows an accuracy advantage when the window size is greater than 3 and demonstrates advantages in time cost and pass rate when the window size is greater than 4.

Moreover, we present the relationship between the maximum window length and the total tracking time in the variable-length window method, as shown in Fig.~\ref{fig:track_time}.
It can be seen that this method has a time advantage compared to the non-tracklet method, but this advantage does not increase with the maximum window length.
This is because the method adaptively controls the scale of the optimization problem during tracklet generation, achieving a balance between tracklet accuracy and generation efficiency.

\subsubsection{Effect of Tracklet Representation}

\begin{figure}[t]
    \includegraphics[width=8.5 cm]{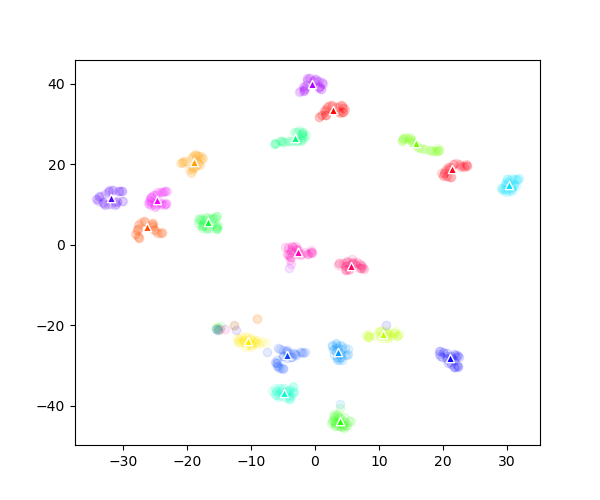}
    \caption{
        Illustration of tracklet feature embedding.
        \label{fig:embedding}
    }
\end{figure}

To clearly understand the differences between independent target representation and tracklet-based representation, we selected segments from the tracking sequence `airplane-0' and visualized the detection features and tracklet features within these segments using the t-SNE method~\cite{van2008visualizing}.
Figure~\ref{fig:embedding} shows the feature distribution of the target's appearance features in a subsequence within the 2D embedding space.
Here, discrete detection target features are represented by dots, while tracklet representations are marked as solid triangles.
The color indicates the target's identity, with the same color representing the same tracklet.
From this figure, it is evident that our tracklet representation is more distinguishable, with clearer feature boundaries and differences compared to discrete detections.

% Figure2_5_01
\begin{figure}[t]
    \includegraphics[width=8.5 cm]{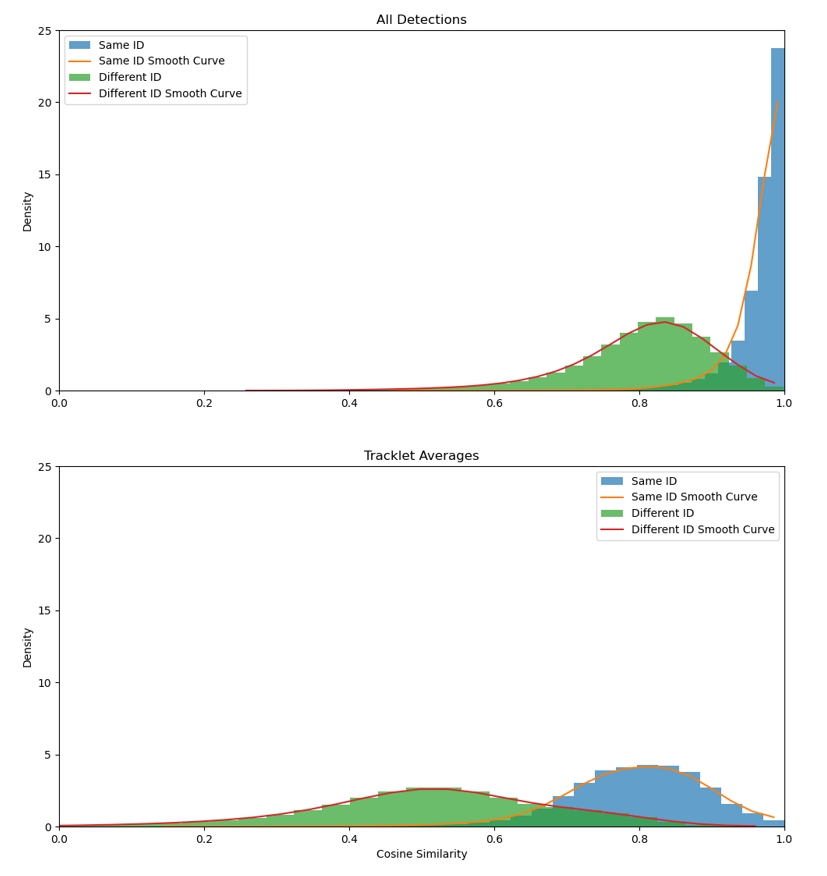}
    \caption{
        Illustration of histograms under discrete feature and tracklet feature.
        \label{fig:histograms}
    }
\end{figure}

In addition, we calculated the similarity based on discrete detection representation and tracklet representation, displaying the differences in histograms, as shown in Figure~\ref{fig:histograms}.
The comparison between the two figures conveys several clear messages: i) In the GMOT task, the similarity between different targets is quite high, mostly above 0.8, posing significant challenges for target association. ii) Compared to independent detection features, tracklet-based feature representation is more discriminative.

\begin{table*}[h]
    \resizebox{\textwidth}{!}{%
        \begin{tabular}{l|cccccccccccc}
            % \hline
            % GT Detection                    &                &                &                &                &                &                &               &              &                &                &                &                 \\
            \hline
            Method                          & MOTA$\uparrow$ & IDF1$\uparrow$ & IDP$\uparrow$  & IDR$\uparrow$  & Rcll$\uparrow$ & Prcn$\uparrow$ & MT$\uparrow$  & PT$\uparrow$ & ML$\downarrow$ & FP$\downarrow$ & FN$\downarrow$ & IDs$\downarrow$ \\
            \hline
            MDP~\cite{xiang2015learning}    & 75.00          & 72.50          & 79.50          & 66.70          & 80.70          & 96.20          & 1105          & \textbf{703} & 136            & 8234           & 49448          & 4103            \\
            DeepSORT~\cite{wojke2017simple} & 80.60          & 79.30          & 85.30          & 74.00          & 84.50          & 97.30          & 1344          & \underline{344}          & 256            & 5944           & 39648          & 4074            \\
            IOU \cite{bochinski2017high}    & 75.90          & 79.00          & 85.80          & 73.20          & 80.40          & 94.20          & 1237          & 260          & 447            & 12704          & 50232          & \textbf{1225}   \\
            FAMNet~\cite{chu2019famnet}     & 67.40          & 70.50          & 86.30          & 60.50          & 70.10          & 97.60          & 1302          & 319          & 323            & 4505           & 76706          & 2454            \\

            SORT~\cite{bewley2016simple}     & 92.43          & 86.54          & \underline{89.55}          & 83.73          & 93.36          & \textbf{99.84} & 1500          & 332          & 110            & \textbf{375}   & {16939}        & 2008            \\
            MHT                             & \textbf{96.37} & \textbf{90.31} & \textbf{89.87} & \textbf{90.76} & \textbf{99.04} & {98.07}        & \textbf{1830} & {73}         & \textbf{39}    & {4982}         & \textbf{2462}  & \underline{1815}          \\
            % \textbf{MHT-APP(Ours)}            & \textbf{97.05} & \textbf{90.71} & \textbf{90.41} & \textbf{91.00} & \textbf{99.16} & 98.511          & \textbf{1828} & 72           & \textbf{42}    & 24633          & 22955          & \textbf{1568}   \\
            {MTT $ow$ $\mathcal{C}$ (Ours)}                         & {90.88}        & {82.95}        & {85.51}        & {80.54}        & {93.07}        & 98.82          & {1509}        & 343          & {90}           & 2835           & 17676          & {2763}          \\
            {MTT(Ours)}                     & \underline{95.60}        & \underline{88.29}        & {89.15}        & \underline{87.45}        & \underline{97.23}        & \underline{99.11}          & \underline{1709}        & 171          & \underline{62}           & \underline{2220}           & 7075           & {1939}          \\

            \hline
        \end{tabular}%
    }
    \caption{Comparison of trackers with GT detection.}
    \label{tab:track_GT}
\end{table*}

\subsection{Ablation Study}
We first defined components $\mathcal{A}$, $\mathcal{B}$, and $\mathcal{C}$.
Component $\mathcal{A}$ represents the appearance feature module, $\mathcal{B}$ represents the tracklet module, and $\mathcal{C}$ represents the adaptive detection clustering method.
To fully demonstrate the effectiveness of the proposed method, we used our method as a baseline and progressively reduced features for comparison.
In this manner, we obtained the corresponding tracking results on the GMOT task (one-shot protocol), as shown in Table~\ref{tab:ablation}.
The experimental results are clear: removing any feature results in lower values for all major tracking metrics.
This demonstrates the effectiveness of our method.

\begin{table}[h]
    \caption{Ablation Study} \label{tab:ablation}
    \begin{tabular}{ccccc}
        \toprule
        Method             & MOTA  & IDF1  & IDs   & FP+FN   \\
        \hline
        Base  & \bf{24.20} & \bf{36.14} & \bf{1,502} & \bf{191,896} \\
        Base - $\mathcal{C}$          & 21.12 & 34.13 & 2,205 & 199,070 \\
        Base - $\mathcal{C}$ - $\mathcal{B}$      & 20.22 & 31.92 & 2,696 & 200,862 \\
        Base - $\mathcal{C}$ - $\mathcal{B}$ - $\mathcal{A}$  & 19.25 & 30.84 & 3,013 & 203,026 \\
        \bottomrule
    \end{tabular}

    % \noindent{\footnotesize{* Tables may have a footer.}}
\end{table}

To evaluate the performance of our method with high-quality input, we compared it with classic methods such as SORT~\cite{bewley2016simple} and DeepSORT~\cite{wojke2017simple} under Ground Truth~(GT) detection, as shown in Table~\ref{tab:track_GT}.
Most baseline algorithms perform well with GT support, with half exceeding a MOTA of 80 and all achieving an IDF1 above 70.
This indicates that with good detection results, even simple tracking strategies can achieve excellent performance.
It aligns with the difficulty of the GMOT task: the lack of robustness of the detector is a major factor in the poor performance of the tracker on this task.

Although the majority of methods perform well, some interesting phenomena can still be observed from the results.
For example, learning-based methods such as FAMNet~\cite{chu2019famnet} and MDP~\cite{xiang2015learning} did not outperform general methods like SORT~\cite{bewley2016simple}, and even DeepSORT~\cite{wojke2017simple} only achieved mid-level performance.
This may partly reflect that the lack of prior information affects the generality of these methods.

Besides, we compared MHT with our MTT method.
Here, MTT $wo$ $\mathcal{C}$ represents the fixed-length window method.
The MHT method has a clear advantage, achieving the best performance under the current settings.
Although our method also performed well, it did not surpass our reproduced MHT method.
It can be said that the segmentation of tracklets somewhat affects the completeness of information during tracking.
Compared to the fixed-length window tracklet method, our adaptive variable-length window method has a smaller impact.

\begin{table*}[h]
    % \resizebox{\textwidth}{!}{%
        \begin{tabular}{l|l|cccccccc}
            \hline
                                        &    Method           & MOTA$\uparrow$ & IDF1$\uparrow$ & MT$\uparrow$ & PT$\uparrow$ & ML$\downarrow$ & FP$\downarrow$ & FN$\downarrow$  & IDs$\downarrow$                                             \\
            \hline
            \multirow{7}*{\shortstack{Normal Detection:\\ Siamese-DETR} }    & IOU~\cite{bochinski2017high}        & 30.7           & 35.0           & 361          &              & 759            & \underline{42504}          & 127024          & 8158                                                        \\
                                        & SORT~\cite{bewley2016simple}        & 30.1           & 34.7           & 235          & -            & {943}          & \bf{29518}     & 145613          & \underline{4060}                                                      \\
                                        & DeepSORT~\cite{wojke2017simple}     & {31.1}         & 41.8           & 382          & -            & 773            & 47336          & 124257          & 5131                                                        \\
                                        & ByteTrack~\cite{zhang2022bytetrack} & {33.7}         & 41.4           & 331          & -            & 764            & 53417          & \underline{104765}          & 4204                                                        \\
                                        & BoT-SORT~\cite{aharon2022bot}       & {34.1}         & \underline{47.5}           & 431          & -            & 674            & 45769          & 119288          & 6775                                                        \\
                                        & TbQ~\cite{liu2024siamese}           & \bf{35.9}      & 42.8           & \underline{504}          &              & \underline{666}            & 44882          & 107894          & 11664                                                       \\
                                        & MTT~(Ours)   & \underline{35.27}         & \bf{48.36}          & \bf{558}          & 769             & \bf{615}            & 67327          & \bf{93844}      & \bf{3998} \\

            \hline
            \multirow{9}*{\shortstack{One-shot Detection:\\ GlobalTrack} }            & MDP~\cite{xiang2015learning}        & {19.80}        & {31.30}        & 142          & \underline{621}          & 1161           & 17260          & 186580          & 1779                                                        \\
                                        & DeepSORT~\cite{wojke2017simple}     & 14.50          & 24.40          & 72           & 509          & 1363           & \underline{9000}           & 208818          & \underline{1315}                                                        \\
                                        & IOU~\cite{bochinski2017high}        & 11.80          & 20.30          & 56           & 397          & 1491           & \textbf{8299}  & 216921          & \textbf{754}                                                \\
                                        & FAMNet~\cite{chu2019famnet}         & 18.00          & 28.30          & 166          & 581          & 1197           & 20741          & 187730          & 1660                                                        \\

                                        & SORT~\cite{bewley2016simple}        & 19.56          & 30.27          & 138          & 615          & 1189           & 30757          & 204168          & 1774                                                        \\
            % SORT++         & 19.82         & 30.59         & 62.37         & 19.98          & 26.145       & 81.61         & 138          & 615          & 1189           & 30757          & 204168          & 1774            \\
                                        & ByteTrack~\cite{zhang2022bytetrack} & 19.10          & 32.10          & 178          & -            & 1069           & 23881          & 181829          & 1791                                                        \\
                                        & BoT-SORT~\cite{aharon2022bot}       & 19.40          & \underline{34.00}          & \bf{251}     & -            & \bf{978}       & 22229          & \underline{176991}          & 7375                                                        \\
                                        & TbQ~\cite{liu2024siamese}           & \underline{20.60}          & 27.40          & 213          & -            & 1066           & 13507          & 182376          & 6407                                                        \\
            % \textbf{MHT with app(Ours)}            & {20.22} & {31.92} & 54.626         & {22.548} & {31.277} & 75.774         & \textbf{227} & \textbf{710} & {1005}  & 25514          & {175348} & 2696            \\
            % \textbf{MHT-tl(Ours)}            & {21.55} & {33.58} & 60.102         & \textbf{30.56} & {31.277} & 78.82         & {210} & {680} & {1052}  & 20947          & {177182} & 2041            \\
                                        & {MTT~(Ours)}                        & \textbf{24.20} & \textbf{36.14} & \underline{231}        & \bf{646}     & \underline{1065}         & 16309          & \textbf{175587} & {1502 }                                                     \\

            \hline
            \multirow{7}*{\shortstack{Zero-shot Detection:\\ GLIP-T} } & IOU~\cite{bochinski2017high}        & 25.10          & 39.30          & 458          & -            & 721            & 55802          & 139560          & 6320                                                        \\
                                        & SORT~\cite{bewley2016simple}        & 25.20          & 40.80          & \underline{354}          & -            & {877}          & 54891          & 143623          & {2987}                                                      \\
                                        & DeepSORT~\cite{wojke2017simple}     & {25.50}        & 41.80          & 401          & -            & 877            & 49984          & 141330          & 2892                                                        \\
                                        & ByteTrack~\cite{zhang2022bytetrack} & {27.00}        & \underline{45.10}          & 447          & -            & 746            & 52591          & \bf{131759}     & \underline{2706}                                                        \\
                                        & BoT-SORT~\cite{aharon2022bot}       & {27.30}        & \bf{49.10}     & 553          & -            & \underline{643}            & 51308          & \underline{133462}          & 4675                                                        \\
                                        & TbQ~\cite{liu2024siamese}           & \underline{27.50}        & 39.80          & 581          &              & \bf{592}       & \underline{49470}          & 136602          & 9972                                                        \\
                                        & MTT~(Ours)                          & \bf{28.11}     & {41.82}        & \bf{314}     & 550          & {1078}         & \bf{28839}     & {152697}        & \bf{1894}                                                   \\

            \hline
        \end{tabular}%
    % }
    \caption{Comparison of trackers with three GMOT evaluation protocols.}
    \label{tab:track_GMOT}
\end{table*}

\subsection{Comparison with Benchmark}

% 另外，需要说明的是，MHT算法有着众多的参数，我们没有动力去找个某个数据集上的最好参数。
We conducted extensive experiments on the GMOT-40 benchmark.
It is important to note that we evaluated all data, including 40 complete sequences.
Table.~\ref{tab:track_GMOT} shows the results of a series of classic and advanced methods on this benchmark.
The evaluation results under three different protocols are included: normal detection, one-shot detection, and zero-shot detection.
Bolded numbers indicate the best results, while underlined numbers represent the second-best results.
Our method achieved good results across multiple metrics and protocols, frequently ranking in the first and second positions.
Specifically, for one-shot detection, we achieved significant improvements of 3.8\% in MOTA and 2.14\% in IDF1 over the second place.
Overall, extensive comparisons demonstrate the potential of our method in the GMOT task.

% \paragraph{Other Dataset: MOT17}
% ....

\section{Limitations and Discussion}\label{sec:limitations}

% \paragraph{Limitation}
Our method has achieved excellent results compared to many others, but it still has some notable limitations.
First, the current tracking performance (below 40 in MOTA and below 50 in IDF1) is still insufficient for the GMOT task.
Detailed tracking results shows that most scenarios exhibit excessive false and missed detections.
This suggests that relying solely on the data association strategy is inadequate, and we have to enhance the robustness of the detector by improving its performance.
Moreover, challenges arise from irregular high-speed dynamic targets, such as the swarm in the `insect-3' sequence of GMOT40, where over-reliance on a single motion model can lead to poor tracking performance.

% \paragraph{Discuss}
The tracklet method reduces many potential assumptions, thereby decreasing the computational complexity of the multi-hypothesis framework.
However, in complex environments, it is still necessary to further consider the efficiency and reliability of the pruning method.
We adopted the same pruning scheme as the MHT framework and believe some possible improvements include:

i) Introducing constraints for creating new trees.
Creating each current tracklet as a new tree undoubtedly increases computational redundancy. Therefore, it is necessary to introduce conditions for creating new trees.

ii) Accelerating the initiation of pruning.
The N-Scan method requires accumulating a certain depth before starting pruning for the first time.
As a result, there may be situations where the computational demand has already exceeded expectations before pruning starts.

iii) Improving pruning flexibility.
Since the existing scheme has a fixed pruning depth, ignoring the requirement for flexibility in cases where some trajectories are highly uncertain can lead to errors during pruning.
Therefore, it is necessary to introduce new strategies to flexibly prune according to different situations.

\section{Conclusion}\label{sec:conclusions}

This paper focuses on general object tracking scenarios where detectors face significant challenges in providing accurate detection results.
In such environments, trackers cannot fully rely on detection results.
Therefore, this paper proposes a variant multi-hypothesis tracking framework and introduces an adaptive tracklet clustering method with variable lengths.
This approach allows for more flexible tracking based on measurement reports, significantly improving tracking efficiency and showing competitive results on GMOT benchmark.

We believe that the practical implications of this work are not limited to tracking methods for low-confidence sensor inputs but may also serve as a parallel tracking preprocessing solution for large-scale group tracking.
We hope to continue research based on this foundation in the future.

% 窗口分类带来的额外的并行能力成为了未来可以考虑的研究之一

% 总结这篇文章的实验结果，贡献。以及不足之处，并展望未来的工作

% \

% Numbered list
% Use the style of numbering in square brackets.
% If nothing is used, default style will be taken.
%\begin{enumerate}[a)]
%\item
%\item
%\item
%\end{enumerate}

% Unnumbered list
%\begin{itemize}
%\item
%\item
%\item
%\end{itemize}

% Description list
%\begin{description}
%\item[]
%\item[]
%\item[]
%\end{description}

%% Figure
%\begin{figure}[<options>]
%	\centering
%		\includegraphics[<options>]{}
%	  \caption{}\label{fig1}
%\end{figure}
%
%
%\begin{table}[<options>]
%\caption{}\label{tbl1}
%\begin{tabular*}{\tblwidth}{@{}LL@{}}
%\toprule
%  &  \\ % Table header row
%\midrule
% & \\
% & \\
% & \\
% & \\
%\bottomrule
%\end{tabular*}
%\end{table}
%
%% Uncomment and use as the case may be
%%\begin{theorem}
%%\end{theorem}
%
%% Uncomment and use as the case may be
%%\begin{lemma}
%%\end{lemma}
%
%%% The Appendices part is started with the command \appendix;
%%% appendix sections are then done as normal sections
%%% \appendix
%
%\section{}\label{}
%
%% To print the credit authorship contribution details
%\printcredits
%
%%% Loading bibliography style file
% \bibliographystyle{model1-num-names}
% \bibliographystyle{cas-model2-names}
\bibliographystyle{elsarticle-num}
%
%% Loading bibliography database
\bibliography{cas-refs.bib}
%
%% Biography
%\bio{}
%% Here goes the biography details.
%\endbio
%
%\bio{pic1}
%% Here goes the biography details.
%\endbio

\end{document}